\documentclass[11pt,peerreview,draftclsnofoot,onecolumn,a4paper]{IEEEtran}
\usepackage{amsmath,amssymb}
\usepackage{microtype}
\usepackage{algorithm,algpseudocode}
\usepackage{graphicx,epsfig,url}
\usepackage[caption=false,font=footnotesize]{subfig}
\usepackage{color}

\usepackage{multirow}

\def\Z{\mathbb{Z}}

\begin{document}

\title{Pruned Non-Local Means}
\author{Sanjay Ghosh, Amit K. Mandal, and Kunal N. Chaudhury\thanks{Correspondence: kunal@ee.iisc.ernet.in. Department of Electrical Engineering, Indian Institute of Science.}}



\maketitle 

\begin{abstract}
 In Non-Local Means (NLM), each pixel is denoised by performing a weighted averaging of its neighboring pixels, where the weights are computed using image patches.
We demonstrate that the denoising performance of NLM can be improved by pruning the neighboring pixels, namely, by rejecting neighboring pixels whose weights are below a certain threshold $\lambda$. While pruning can potentially reduce pixel averaging in uniform-intensity regions, we demonstrate that there is generally an overall improvement in the denoising performance. In particular, the improvement comes from pixels situated close to edges and corners.  The success of the proposed method strongly depends on the choice of the global threshold $\lambda$, which in turn depends on the noise level and the image characteristics. We show how Stein's unbiased estimator of the mean-squared error can be used to optimally tune $\lambda$, at a marginal computational overhead. We present some representative denoising results to demonstrate the superior performance of the proposed method over NLM and  its variants.
\end{abstract}

\section{Introduction}

We consider the standard problem of denoising grayscale images that are corrupted with additive white Gaussian noise. 
The setting here is that we are given a \textit{noisy} image
\begin{equation}
\label{noise}
y_i = x_i  +  n_i \qquad (i \in I),
\end{equation}
where $I$ is some finite rectangular domain of $\Z^2$, $\{x_i : i \in I\}$ is the unknown \textit{clean} image, and $\{n_i : i \in I\}$ are i.i.d. $\mathcal{N}(0,\sigma^2)$ where $\sigma$ is the standard deviation of the Gaussian noise. The focus of this work is the Non-Local Means (NLM) algorithm that was introduced by Buades, Coll, and Morel in \cite{BCM2005cvpr,BCM2005}. NLM denoises each pixel by taking a weighted average of the neighboring pixels. A key innovation in NLM is that the weights are computed using patches (blocks of neighboring pixels) instead of pixels.  More specifically, the \textit{denoised} image $\{\hat{x}_i : i \in I\}$ is computed using the formula 
\begin{equation}
\label{NLM} 
\hat{x}_i =  \frac{\sum_{j \in S(i)} w_{i,j} y_j}{\sum_{j \in S(i)} w_{i,j} },
\end{equation}
where $S(i)$ denotes the \textit{neighborhood} around pixel $i$. The original proposal in \cite{BCM2005cvpr} was to set $S(i)$ to be the whole image. In practice, one restricts $S(i)$ to a sufficiently large window of size $(2S +1) \times (2S+1)$ centered at $i$, that is, $S(i) = i +[-S,S]^2$. The weights $\{w_{i,j}: i \in I, \ j \in S(i)\}$ are set to be
\begin{equation}
\label{weights}
 w_{i,j} =  \text{exp} \Big(\! - \frac{1}{h^2} \sum\limits_{k \in \mathcal{P} } (y_{i+k} - y_{j+k})^2 \Big),
\end{equation}
where $h$ is a smoothing parameter, and \textit{patch} $\mathcal{P} = [-K, K]^2$ is a square neighborhood of the origin of size $(2K +1) \times (2K+1)$. 

Since its introduction, several variations of NLM have been proposed which can further improve the denoising. For example, various weighting schemes for the center pixel have been proposed in \cite{S2010,DAG2011a,WTNN2013a}. A weighting scheme based on quadratic programming was proposed in \cite{FSHYZ2015}, which exploits the overlapping information between adjacent patches. Further, optimization framework based approaches have been proposed in \cite{ZFW2013, LZJ2015} to improve the performance of NLM. In \cite{RZ2011}, the weighting was performed based on the structural similarity indices between patches. In a different direction, it was demonstrated in \cite{OEW2008,T2009,VK2011,Z2016} that the performance (and the speed) of NLM can be improved by  computing the weights from low-dimensional projections of the patches. 

{
The primary application of NLM is noise reduction \cite{Tracey2012, Bora2015}. However, since the inception of NLM, more sophisticated algorithms with superior denoising performances have been proposed  \cite{KSVD2006, BM3D2007, Milanfar2014, Duan2015, Yue2015, RE2015, Dinh2016, Yang2011, Yang2012}. While NLM and its variants are no longer the state-of-the-art in image denoising, they nevertheless continue to be of interest due to their simplicity, decent denoising performance, and the availability of fast implementations.
We also note that several researchers have used NLM (either directly or via post-processing) in image restoration applications such as deblurring \cite{Chen2010, Jung2011}, inpainting \cite{Wong2008}, demosaicking \cite{Buades2009}, and super-resolution \cite{Protter2009, Zhang2012, Zhang 2015}.}

The investigation in this paper was particularly motivated by the work on shape-adaptive patches \cite{DDS2012} and robust $\ell_1$ regression \cite{CS2012,SC2013}. In either case, it was demonstrated that the denoising quality of NLM can be improved by adapting the averaging neighborhood at every pixel. While this was done explicitly in \cite{DDS2012} in terms of the patch shape, the adaptation was achieved implicitly in \cite{CS2012,SC2013} via an optimization. In this paper, we propose to perform the adaptation by explicitly pruning those neighboring pixels whose weights are below a certain global threshold $\lambda$. We use Stein's unbiased estimator of the mean-squared error to optimally tune $\lambda$. In other words, we optimize the denoising quality by performing a simple one-dimensional optimization for the whole image. As a result, the overall implementation is much faster than {those implementations in \cite{DDS2012,CS2012,SC2013,CS2013}}, where one is required to perform an adaptation at every pixel.

The rest of the paper is organized as follows. In Section \ref{S2}, we motivate the main idea using a simple experiment, and then present the proposed denoising algorithm and the implementation details. The denoising results are reported in Section \ref{S3}, where we also compare the proposed method with NLM and its variants.

\section{Pruned Non-Local Means}
\label{S2}

We now perform some numerical experiments {to reflect the motivation that pruning can improve} the denoising performance. We consider a scenario where the pixel of interest (POI) is situated closed to a edge, as shown in the example in Figure 1(a). We compute the NLM weights around the POI (marked with a solid black circle) using \eqref{weights}. We expect the weights of neighboring pixels that are on the same side of the edge as the POI to be generally higher than {the weights of pixels} that are on the other side of the edge. As shown in Figure 1(b) this is indeed the case if the weights are computed from the (unknown) clean edge. Unfortunately, this is not the case if the weights are computed from the noisy edge. As seen in Figure 1(b), the weights of some of the pixels on the left of the edge (same side of the edge as the POI) are smaller than the weights of some of the pixels on the right. This leads to a poor averaging in \eqref{NLM} -- pixels from both sides of the edge are mixed. For example, in Figure 1(a), the intensities at the POI before and after adding noise are $0$ and $-3.75$. Using NLM, the denoised output is $81.72$ (as a reference, the NLM output using the weights computed from the clean patches is 73.09). On the other hand, notice that if we sort the noisy weights in Figure 1(b), then the majority of the pixels on the left of the edge would have the largest weights. Indeed, if we use the top $50\%$ neighbors (neighbors with largest weights) for the averaging, then the output comes down to $42.32$ (which is even better than that obtained using the clean weights). This suggests a means of improving the quality of denoising, namely, by restricting the averaging in \eqref{NLM} to those neighboring pixels that are assigned top weights.
\begin{figure}
  \centering
  \subfloat[clean and noisy edges ($\sigma=80$).]{\label{fig1a}\includegraphics[width=0.5\linewidth]{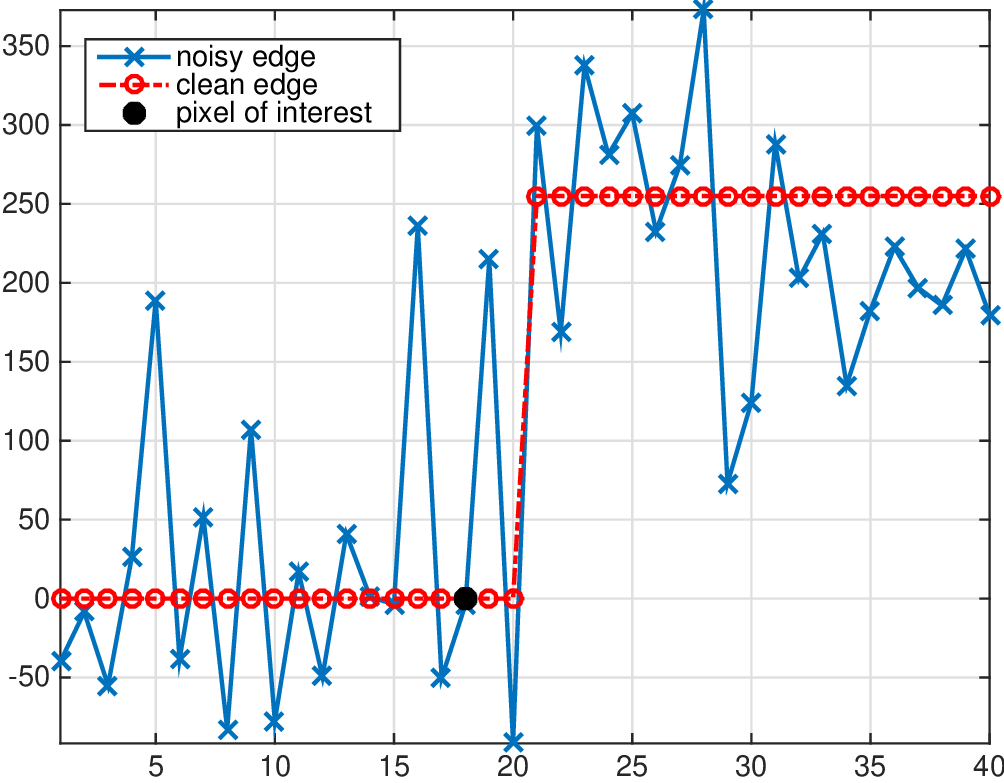}} 
  \subfloat[neighboring weights for the POI.]{\label{fig1b}\includegraphics[width=0.5\linewidth]{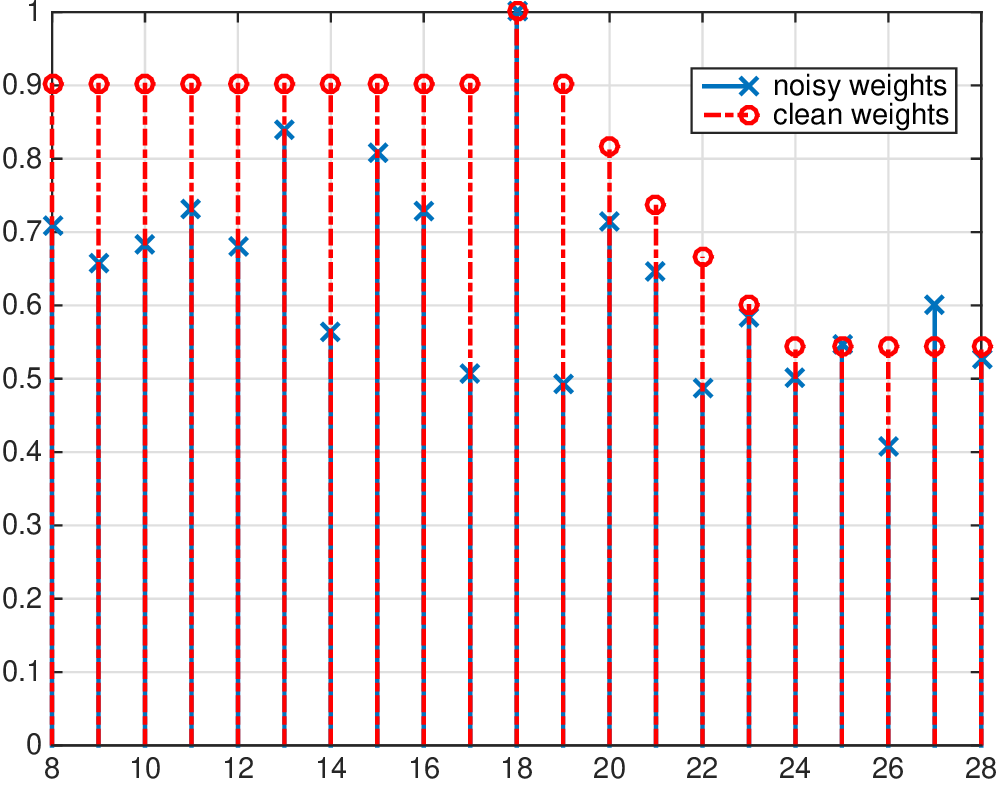}}
  \caption{NLM-based denoising at a pixel (index = $18$) situated close to an ideal edge. NLM parameters: $S = 10, K = 3$, and $h = 10 \sigma$.}
\end{figure}
Based on the above observation, we propose to perform \textit{pruning} as follows. We fix a threshold $0 < \lambda < 1$, and consider the step function $\phi_{\lambda} : [0,1] \rightarrow [0,1]$ given by
\begin{equation}
\label{phi}
\phi_{\lambda}(t) = 
   \begin{cases}
    1, \ & \text{for} \ \lambda \leq t \leq 1  \\
    0, & \text{for} \ 0 \leq t < \lambda.
  \end{cases}
\end{equation}
The process of nullifying the weights that are smaller than $\lambda$ can be achieved by {modifying \eqref{noise} as}:
\begin{equation}
 \label{hardPNLM} 
 \hat{x}_i = \frac{\sum_{j \in S(i)}  \ w_{i, j} \  \phi_{\lambda}(w_{i, j}) \ y_j} { \sum_{j \in S(i)} w_{i, j}  \ \phi_{\lambda}(w_{i, j}) }. 
\end{equation}
To avoid introducing unnecessary symbols, we continue to use the same notation for the denoised image as in \eqref{NLM}. {Notice that the weights defined in \eqref{weights} take values in $[0,1]$; therefore, the variable $t \in [0, 1]$ in \eqref{phi}.} When $\lambda=0$, none of the weights are pruned, and when $\lambda$ is close to unity, a large fraction of the weights are pruned. The optimal setting would be somewhere in between. We can ideally tune $\lambda$ by minimizing the mean-squared-error (MSE) given by
\begin{equation}
\label{MSE}
\text{MSE}= \frac{1}{|I|} \sum_{i \in I} \big(\hat{x}_i - x_i \big)^2,
\end{equation}
{where $|I|$ is the size of the image, i.e., the number of pixels in the image}. However, since we do not have access to the clean image $\{x_i\}$, we cannot compute the MSE in practice. 
 \begin{figure}
 \centering
  \includegraphics[width=0.4\textwidth]{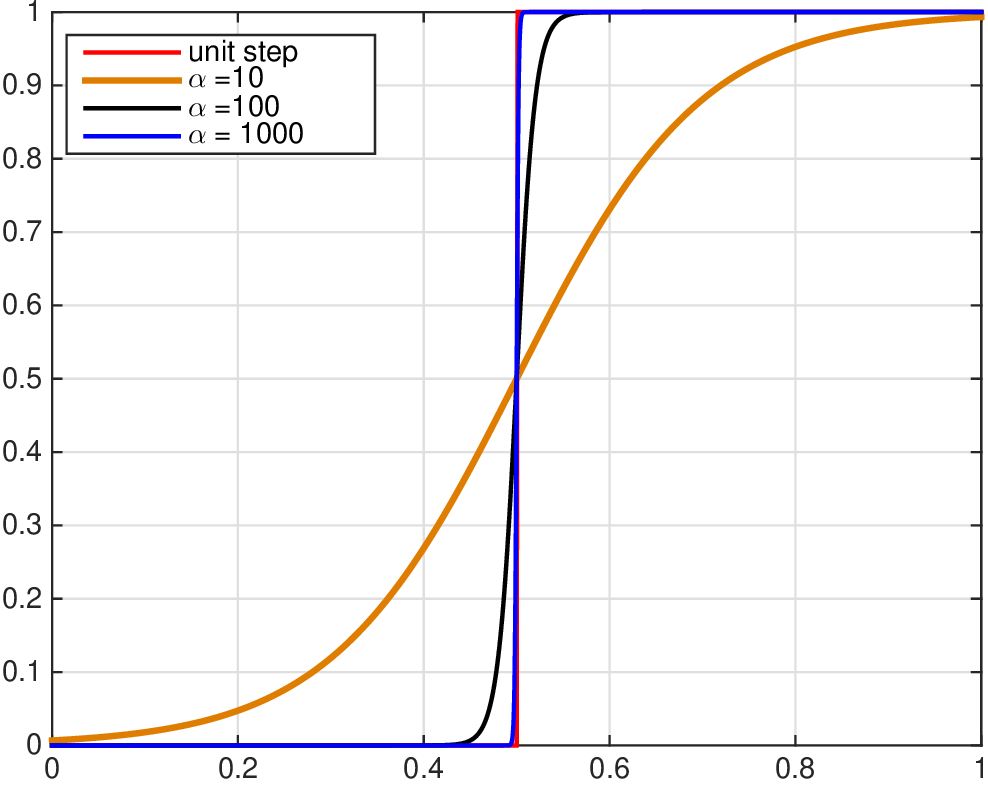}
  \caption{Sigmoid function in \eqref{sigmoid} when $\lambda = 0.5$. At large values of $\alpha$, we obtain a close approximation of the unit step function.} 
  \label{plotSigmoid} 
\end{figure}
This is precisely where Stein's Unbiased Risk Estimator (SURE) proves to be useful. This is given by
\begin{equation}
\label{sure}
\text{SURE} = \frac{1}{| I |} \sum_{i \in I} \big(\hat{x}_i - y_i \big)^2 - \sigma^2 + \frac{2\sigma^2}{|I|} \sum_{i \in I} \frac{\partial \hat{x}_i}{\partial y_i},
\end{equation}
{where we recall that $\sigma$ is the standard deviation of the additive Gaussian noise in \eqref{noise}}.
\begin{figure}
\centering
\subfloat[clean and noisy edges ($\sigma=80$).]{\includegraphics[scale=0.45]{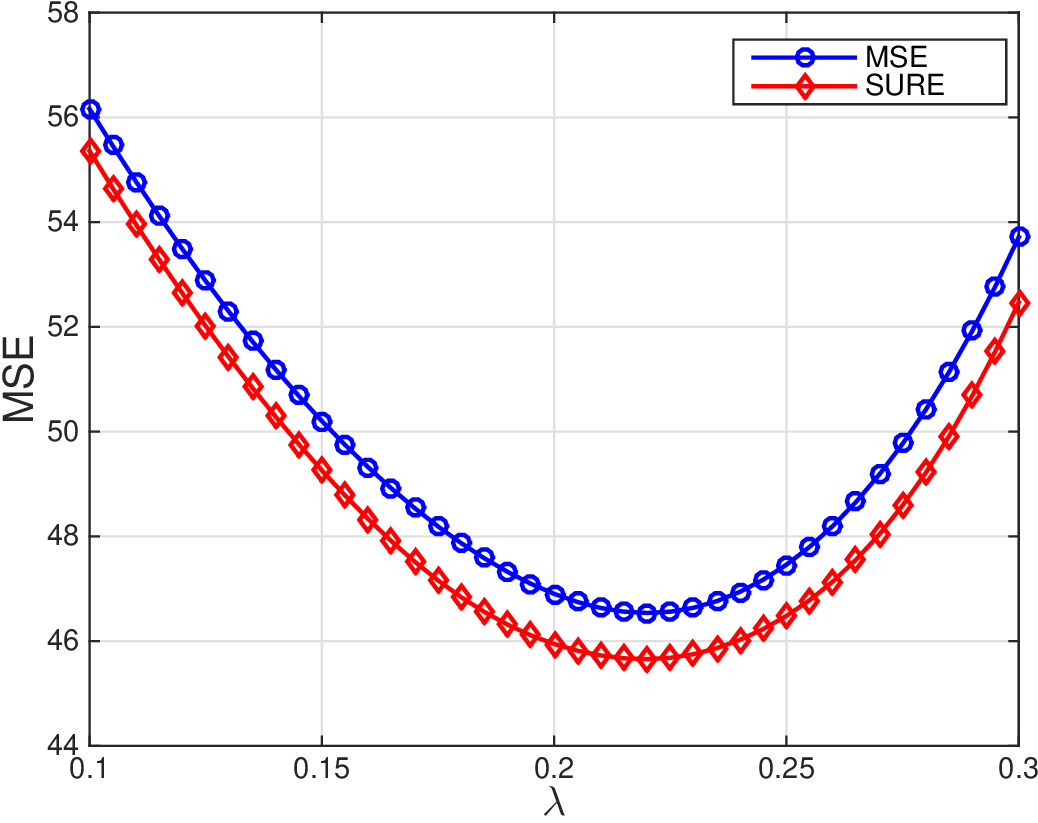} }
\subfloat[clean and noisy edges ($\sigma=80$).]{\includegraphics[scale=0.45]{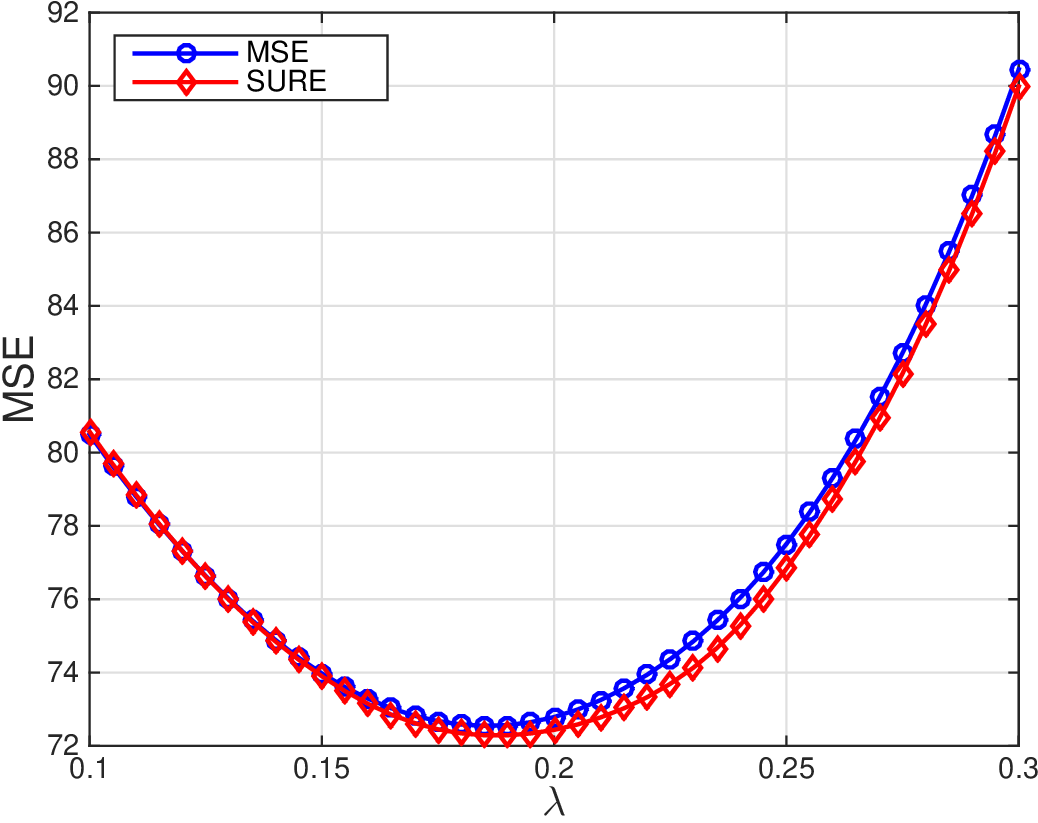}} 
\caption{Comparison of MSE (computed from the clean image) and SURE as a function of $\lambda$ for two different images ($\sigma=20$).}
\label{MSEvsSURE}
\end{figure}
SURE has the remarkable property that its expected value equals that of the MSE for the noise model in \eqref{noise} \cite{S1981}. This makes it an useful surrogate for the MSE -- it can be computed without the knowledge of the clean image. The difficulty in applying SURE directly to the transformation in \eqref{hardPNLM} is that \eqref{phi} is not differentiable and therefore we cannot compute the partial derivatives in \eqref{sure}. This is, however, not a fundamental problem and we can approximate \eqref{phi} by a differentiable function. For example, we propose to use the following sigmoid function:
\begin{equation}
\label{sigmoid}
\phi_{\lambda, \alpha}(t) =     \frac{1}{1 + e^{ - \alpha(t - \lambda)} } \qquad (0 \leq t \leq 1),
\end{equation}
where $\alpha > 0$ is the slope of the sigmoid at $t=\lambda$. {It can be verified that \eqref{sigmoid} converges pointwise to \eqref{phi} as $\alpha$ get large, namely, for fixed $0 \leq t \leq 1$,
\begin{equation*}
\lim_{\alpha \rightarrow \infty} \phi_{\lambda, \alpha}(t) = \phi_{\lambda}(t).
\end{equation*}}
In Figure \ref{plotSigmoid}, we plot \eqref{sigmoid} for different values of $\alpha$ and compare it with \eqref{phi} for the same $\lambda$. Ideally, we should use a  large $\alpha$ to obtain a good approximation. However, as will be evident shortly, the partial derivatives in \eqref{sure} explode as $\alpha$ increases. Following exhaustive simulations, we decided to set $\alpha=100$. The approximation is reasonable in this case and the derivatives can be computed stably. In other words, we consider the following formula 
\begin{equation}
 \label{pnlm} 
 \hat{x}_i = \frac{\sum_{j \in S(i)} w_{i, j}   \phi_{\lambda,\alpha}(w_{i, j}) \ y_j} { \sum_{j \in S(i)} w_{i, j}  \phi_{\lambda,\alpha}(w_{i, j}) },
\end{equation}
where $\alpha=100$. We will refer to \eqref{pnlm} as Pruned Non-Local Means (PNLM). We note that the SURE for \eqref{NLM} was originally derived in \cite{VK2009}, and was shown to be close to the MSE. Similarly, we can derive the SURE for \eqref{pnlm}. In particular, the partial derivative of \eqref{pnlm} is given by (cf. Appendix): 
\begin{align}
\label{sure_pnlm}
&\frac{\partial \hat{x}_i}{\partial y_i} =  \frac{1}{W_i} \Big[\psi(1)  + \frac{2}{h^2} \!\! \sum_{j \in S(i)}  \!\! \! w_{i,j} \ \psi'(w_{i,j}) (y_j - \hat{x_i}) (y_j - y_i)  \nonumber \\
& + \frac{2}{h^2} \! \sum_{k \in \mathcal{P}}  w_{i,i+k}  \ \psi'(w_{i,i+k}) (y_{i+k} - \hat{x_i}) (y_{i- k} -  y_i) \Big],
\end{align}
where 
\begin{equation}
\label{temp}
W_i = \sum_{j \in S(i)} \psi(w_{i,j}) \quad \text{and} \quad \psi(t)=t \phi_{\lambda,\alpha}(t).
\end{equation}
The  expression of $\psi'(t)$ is given in \eqref{psi'} for completeness. A comparison of the MSE and SURE is provided in Figure \ref{MSEvsSURE} when $\alpha=100$. 
We notice that the gap between MSE and SURE decreases with the increase in image size. Importantly, we note that the minima of MSE and SURE are attained at almost the same $\lambda$.
 \begin{figure}
 \centering
  \includegraphics[width=0.6\textwidth]{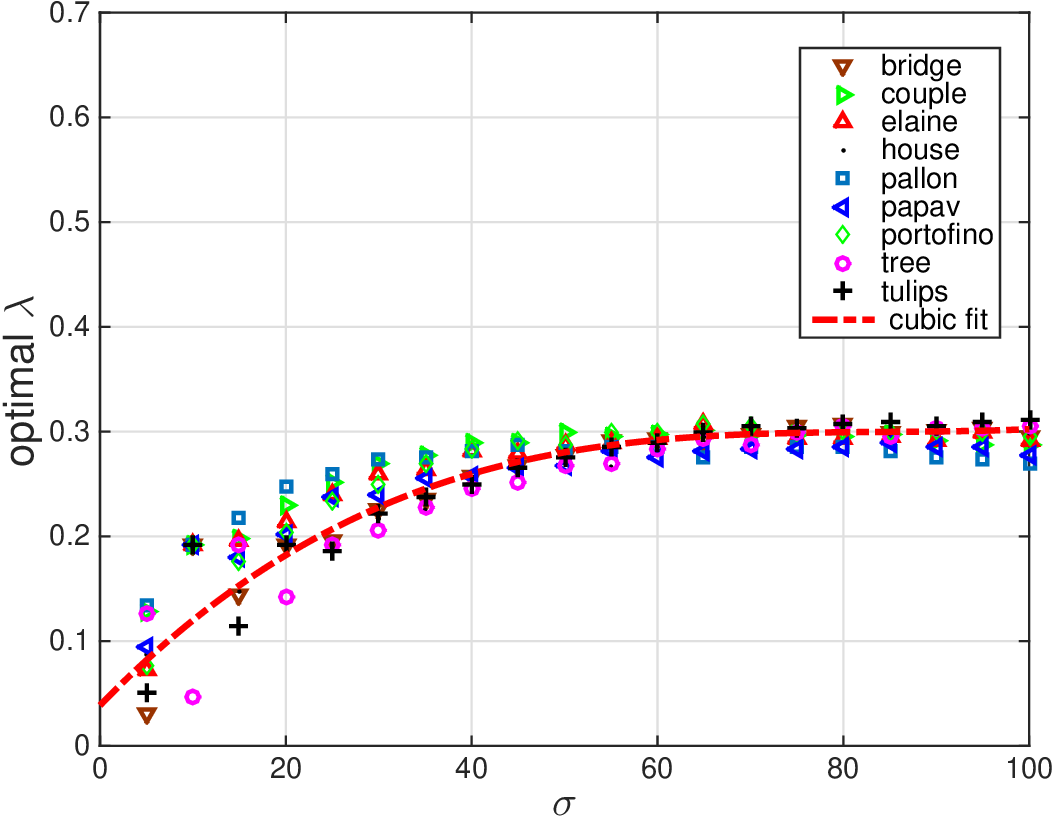}
  \caption{Optimal $\lambda$ versus noise level for various natural images \cite{ImgDatabase1,ImgDatabase2}. The parameters used are $S = 10$, $K = 3$, and $h = 10 \sigma$.}
  \label{fig5}
\end{figure}

\section{Experimental Results}
\label{S3}

\subsection{Parameters}
The denoising performance of PNLM clearly depends on the choice of $\lambda$. In particular, the optimal $\lambda^{\ast}$ (minimizer of MSE) depends on the noise level and the image characteristics. In order to obtain a noise-based rule for setting $\lambda^{\ast}$, we took a large collection of natural images \cite{ImgDatabase1,ImgDatabase2} and computed the optimal $\lambda$ at various noise levels using exhaustive search. The scatter plot of $\sigma$ and the corresponding $\lambda^{\ast}$ for some of the images is shown in Figure \ref{fig5}. Also shown is the cubic polynomial that we obtained using least-square fitting: 
\begin{equation}
\label{cubicFit}
\lambda^{\ast} = 4.3 \times 10^{-7} \sigma^3 - 1.1 \times 10^{-4} \sigma^2 + 9.2  \times 10^{-3}\sigma + 0.039. 
\end{equation}
{We note that the above fitting pertains to the parameter setting $S=10, K=3$ and $h = 10 \sigma$. In this case, we can directly use \eqref{cubicFit} to predict $\lambda^{\ast}$. Of course, \eqref{cubicFit} will not provide a good estimate of $\lambda^{\ast}$ for other parameter settings. One option is to maintain a lookup table of the optimal rules for selected parameter settings, and then interpolate between them. Instead, we propose to use \eqref{cubicFit} to initialize a one-dimensional search, irrespective of the parameter setting. Starting from a coarse estimate of $\lambda^{\ast}$, we then use golden section search \cite{K1953} to refine the estimate}. The complete process is summarized in Algorithm \ref{algo}. Note that, compared to NLM, we have to perform an additional search in PNLM.
Each step of this search requires us to compute \eqref{sure} and \eqref{pnlm} twice. However, the weights in \eqref{sure} and \eqref{pnlm} are computed using the patch distances used in NLM. Therefore, we have to compute the patch distances just once, which happens to be the computation-intensive part of NLM. 
The iterations in Algorithm \ref{algo} are terminated when the change in $\lambda^{\ast}$ between successive iterations is within $10^{-4}$. 
The whole process is quite fast once the patch distances have been computed. 

\textcolor{black}{The denoising performance of PNLM and NLM are compared for different choices of search window and patch size in Table \ref{PNLM:parameterAnalysis}. We see that PNLM achieves better denoising irrespective of the choice of parameters.}
{We note that the optimal selection of NLM parameters is an altogether different line of work \cite{S2010, {VK2011}}. In this paper, the focus is on post-processing of weights to enhance the robustness of NLM. The idea of pruning remain relevant irrespective of the choice of NLM parameters. For the present discussion, we choose the NLM parameters to be $S = 10, K = 3$ and $h = 10 \sigma$, as was originally suggested in \cite{BCM2005cvpr}.  As far as the additional parameters $\alpha$ and $\lambda$ in PNLM are concerned, we set $\alpha = 100$ to ensure a reasonable approximation of the step function $\phi_{\lambda}$, and the crucial parameter $\lambda$ is set using the process described above.}

\begin{table*}
\setlength{\tabcolsep}{4.0pt}
\caption{Comparison of the denoising performance (PSNR/SSIM) for different search window ($S$) and patch size ($K$).
Noise level used is $\sigma = 20$. 
The results are for the \textbf{House} image \cite{ImgDatabase1}.}
\vspace{2mm}
\centering 
\begin{tabular}{|c | c| c | c  |}
\hline
$K$ & $S$  & NLM   & PNLM   \\
\hline
 & 7  	& 28.99 / 80.47   & 32.22 / 84.47  \\
 & 10 	& 28.31 / 79.98   & 31.97 / 84.36  \\ $K = 2$
 & 12 	& 28.03 / 79.86   & 31.79 / 84.25  \\
 & 15 	&27.60 / 79.53   & 31.52 / 84.00  \\
 & 20  	& 26.97 / 78.97   & 31.16 / 83.67  \\
\hline
 &7	& 30.45 / 82.42   & 32.26 / 84.77  \\
 & 10	& 29.78 / 81.84   & 32.20 / 84.97  \\ $K = 3$
 & 12	& 29.45 / 81.59   & 32.19 / 85.06  \\
 & 15	& 28.98 / 81.08   & 32.14 / 85.09  \\
 & 20	& 28.33 / 80.35   & 32.05 / 85.03 \\
\hline
 & 7	& 31.61 / 83.93   & 31.59 / 84.27  \\
 & 10	& 31.21 / 83.60   & 31.73 / 84.53  \\ $K=4$
 & 12	& 30.94 / 83.38   & 31.77 / 84.64  \\
 & 15	& 30.52 / 82.92   & 31.78 / 84.69  \\
 & 20	& 29.86 / 82.16   & 31.75 / 84.61 \\
 \hline
\end{tabular}
\label{PNLM:parameterAnalysis}
\end{table*}

\begin{algorithm}[!htp]
\caption{Golden Section Search}
\label{algo}
\begin{algorithmic}
     \State \textbf{Input}: Noisy image $\{y_i\}$; noise level $\sigma$;  parameters  $S$, $K$, and  $h$.
      \State \textbf{Return}: Estimate of $\lambda^{\ast}$.
      \State \textbf{Initialize}: $k=0$; $\rho = 0.618$; $\delta = 0.05$; set $\lambda_0$ using \eqref{cubicFit}.
      \State {\bf Loop until converge}    
      \State 1. Set $l_k = \lambda_k - \delta$ and $u_k= \lambda_k + \delta$. 
	\State 2.  Set $p_k = u_k - \rho (u_k -l_k)$ and $q_k = l_k + \rho (u_k - l_k)$.
      \State 3.  If $\mathrm{SURE} (p_k) >  \mathrm{SURE} (q_k)$
      \\ \quad  \quad set $l_{k+1} = p_k$ and $u_{k+1} = u_k$.
      \\ \quad { Else} 
      \\ \quad  \quad set $l_{k+1} = l_k$ and $u_{k+1} = q_k$. 
      \\ \quad \ $k = k+1$; \\
{\bf End Loop}     
  \State   return $(l_k + u_k)/2$.
\end{algorithmic}
\end{algorithm}

\subsection{Runtime and Denoising Results}

We now present simulation results concerning the run time of the proposed algorithm.
In Table \ref{table:runtime}, we compare the {computation costs} of NLM and PNLM on various image sizes. We see that the computational overhead for PNLM is within $30\%$ of that of NLM.
In this context, we note that one can use any of the existing fast algorithms for NLM to speedup the PNLM computation \cite{KNC2016, Darbon2008}. By using an optimized C implementation of any of these fast algorithms, one can bring down the run-time to within a second for a $512 \times 512$ image.

 \begin{table*}[!htp]
\setlength{\tabcolsep}{6pt}
\caption{Comparison of the run times of NLM and PNLM for  various image sizes, using the standard parameters $S=10$ and $K=3$. The computations were performed using Matlab 8.4 on a $3.4$ GHz Intel quad-core machine with $32$ GB memory. \vspace{2mm}}  
 \vspace{1mm}
\centering
\begin{tabular}{|c|c|c|c|c|}
\hline
Method       & $64\times 64$ & $128\times 128$  & $256\times 256$ & $512\times 512$  \\
[0.2 cm] \hline
NLM              & 7.7 s   & 30 s  & 122 s    & 490 s   \\ [0.1 cm]
\hline
PNLM           & 9.7 s   & 39 s  & 157 s    & 630 s   \\ [0.1 cm]
\hline
\end{tabular}
\label{table:runtime}
\end{table*}

We now report some representative denoising results. We compared PNLM with NLM and the following variants of NLM: NLEM \cite{CS2012}, PCA-NLM \cite{T2009}, Zero-CPW \cite{DAG2011a}, One-CPW \cite{DDS2012}, and Max-CPW \cite{S2010}.
{For all experiments, we used the following parameter settings: $S=10$, $K=3$ and $h = 10 \sigma$} \cite{BCM2005cvpr}. 
In Table \ref{PSNR_SSIM1} and \ref{PSNR_SSIM3}, we report the PSNR and SSIM \cite{SSIM2004} obtained using various denoising algorithms on the  \textit{Boat} and \textit{Hill} images at different noise values. We use the standard definition
\begin{equation}
 {\text{PSNR}=10 \log_{10}(255^2/\text{MSE})},
\end{equation}
 where MSE is given by \eqref{MSE}.
 
\begin{table*}[!htp]
\setlength{\tabcolsep}{2.80 pt}
\caption{Comparison of PSNR and SSIM with NLM and its variants for the \textbf{Boat} image \cite{ImgDatabase1}. 
The highest PSNRs and SSIMs are marked in bold. \vspace{2mm}}
\centering 
 \begin{tabular}{|c|c|c|c|c|c|c|c|} 
\hline
{  $\sigma$ }  & NLM  & NLEM      & {PCA-NLM } 
& Zero-CPW  & One-CPW  & Max-CPW  & PNLM 
\\  & \cite{BCM2005cvpr} & \cite{CS2012}   & \cite{T2009}  & \cite{DAG2011a}  & \cite{DDS2012} & \cite{S2010} & \\
[0.15 cm]
\hline 
5 & 35.07/96.29 & 35.04/96.89  & 34.94/96.07 & 28.88/85.37 & 29.84/86.34 & 29.28/85.80 & {\bf 35.78/97.90}  \\ [0.02 cm]  \hline
10 & 30.78/88.95 & 30.96/91.11  & 30.71/88.74 & 28.83/85.50 & 29.91/86.88 & 29.23/85.95 & {\bf 32.22/94.37} \\   [0.02 cm] \hline
20 & 26.73/77.96 & 27.43/83.13  & 26.67/77.79 & 28.45/85.58 & 27.83/88.72 & 28.83/86.11 & {\bf 28.95/87.12} \\ [0.02 cm] \hline
30 & 24.72/70.66 & 25.58/77.60 & 24.70/70.55 & 27.04/83.37 & 19.92/65.45 & 27.11/83.70 & {\bf 27.29/81.74} \\ [0.02 cm] \hline
40 & 23.64/65.68 & 24.60/73.47  & 23.62/65.60 & 23.87/74.36 & 16.12/48.78 & 23.29/72.74 & {\bf 26.33/77.63} \\ [0.02 cm] \hline
50 & 22.96/62.13 & 23.99/70.09  & 22.95/62.06 & 20.03/58.08 & 14.15/40.68 & 19.36/56.86 & {\bf25.18/73.92} \\ [0.02 cm]\hline
60 & 22.48/59.45 & 23.48/66.97  & 22.47/59.40 & 16.91/43.81 & 12.57/34.51 & 16.56/45.27 & {\bf24.41/70.43} \\ [0.02 cm] \hline
70 & 22.15/57.43 & 23.03/64.00  & 22.14/57.39 & 14.38/32.53 & 11.23/29.60 & 14.38/35.96 & {\bf23.76/67.07} \\ [0.02 cm]  \hline
80 & 21.87/55.68 & 22.56/60.96  & 21.86/55.66 & 10.95/15.50 & 10.07/25.57 & 11.28/18.87 & {\bf23.16/63.84} \\ [0.02 cm]  \hline
90 & 21.63/54.16 & 22.12/58.03  & 21.64/54.16 & 07.01/03.81 & 09.04/22.28 & 07.25/04.69 & {\bf22.67/60.89} \\ [0.02 cm] \hline
100 & 21.46/52.89 & 21.69/55.23  & 21.47/52.92 & 05.59/01.09 & 08.13/19.65 & 05.65/01.27 & {\bf22.26/58.21}  \\ [0.02 cm]
\hline
\end{tabular}
\label{PSNR_SSIM1}
\end{table*}

\begin{figure}[!htp]
  \centering
   \subfloat[Clean image (256 $\times$ 256).]{\includegraphics[width=0.4 \linewidth]{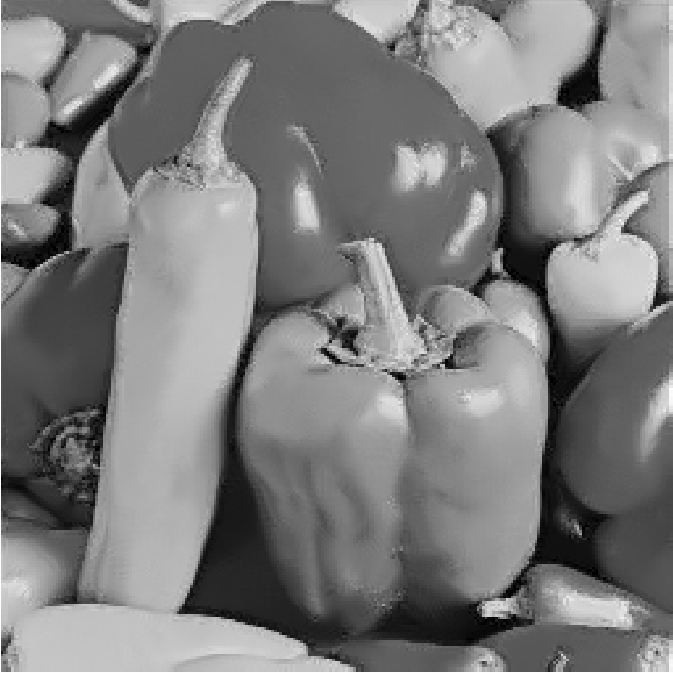}}
 \hspace{0.1mm}
 \subfloat[Noisy image: 28.11 / 67.74.]{\includegraphics[width= 0.4 \linewidth]{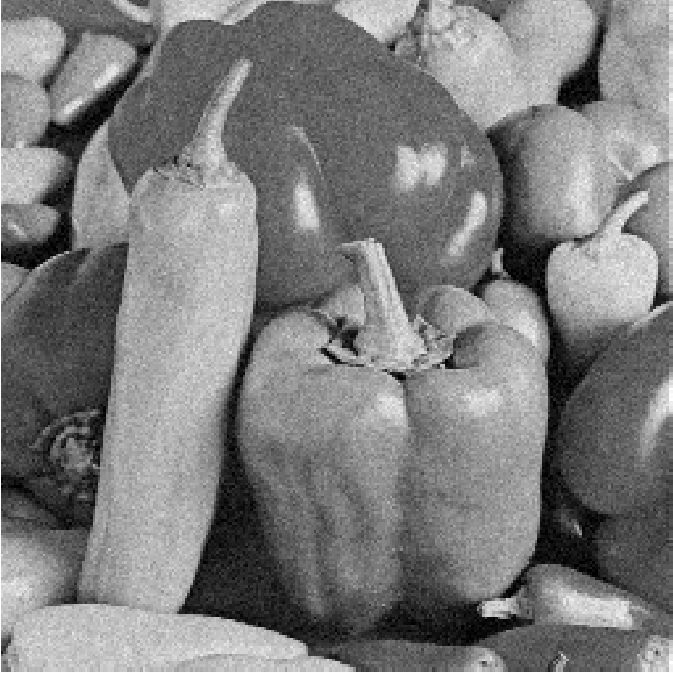}}  \\
\subfloat[NLM \cite{BCM2005cvpr}: 32.29 / 88.86.]{\includegraphics[width= 0.40 \linewidth]{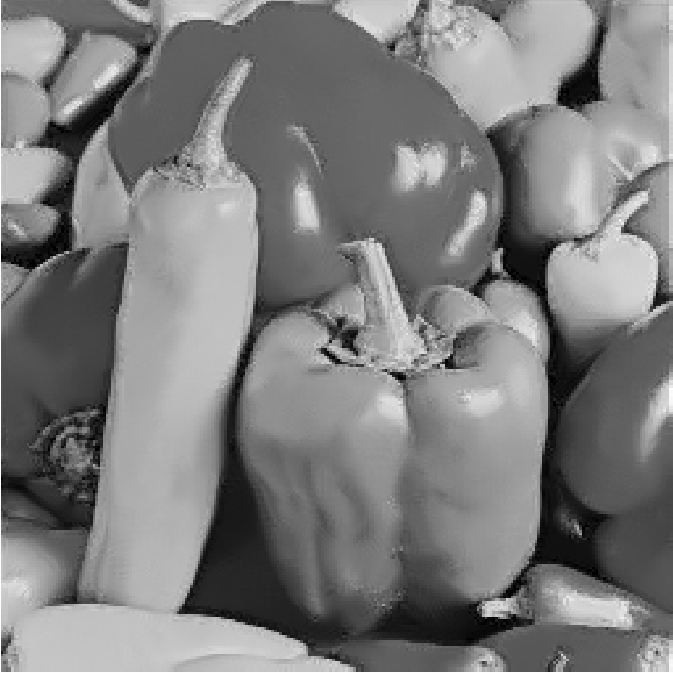}}  \hspace{0.1mm}
  \subfloat[PNLM: {\bf 33.01} / {\bf 90.85}.]{\includegraphics[width= 0.4 \linewidth]{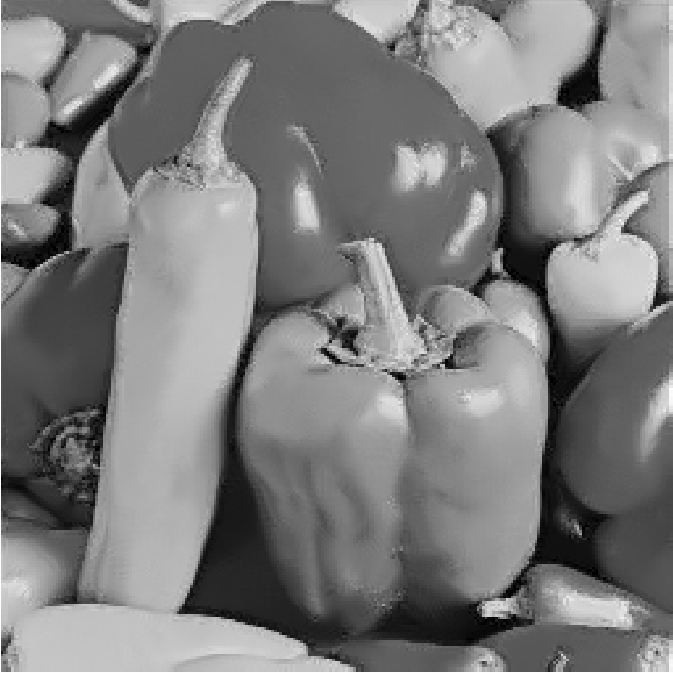}} 
  \caption{Comparison of the denoising results at $\sigma = 10$ for the \textit{Peppers} image \cite{ImgDatabase1}. The corresponding PSNR/SSIM values are shown in the caption.}
    \label{fig:pep_set}
\end{figure}

In Table \ref{PSNR_SSIM2}, we compare the PSNR and SSIM of {various denoising algorithms} on different natural images, with the noise fixed at $\sigma=50$. 
From these results, we see that PNLM consistently outperforms NLM  in terms of PSNR and SSIM at all noise levels. 
For a visual comparison, the denoised images obtained using NLM and PNLM are provided in Figure  \ref{fig:pep_set}.
Notice that the output of PNLM is visibly sharper than that obtained using NLM. 

\begin{table*}[!htp]
\setlength{\tabcolsep}{2.6pt}
\caption{Comparison of PSNR/SSIM with NLM and its variants on the \textbf{Hill} image \cite{ImgDatabase1}. 
The highest PSNRs and SSIMs are marked in bold.\vspace{2mm}} 
\centering 
 \begin{tabular}{|c|c|c|c|c|c|c|c|} 
\hline
{  $\sigma$ }  & NLM  & NLEM     & PCA-NLM & Zero-CPW & One-CPW  & Max-CPW & PNLM
\\ [0.2 cm]
\hline 
5 & 34.90/95.30 & 34.98/96.17  & 34.66/94.99 & 28.86/83.41 & 29.37/84.20 & 29.06/83.76 & {\bf 35.78/98.52}  \\ [0.02 cm]
\hline
10 & 30.36/86.55 & 30.84/89.36 &   30.23/86.29 & 28.87/83.57 & 29.55/84.77 & 29.07/83.93 & {\bf 32.15/94.01} \\ [0.02 cm]
\hline
20 & 26.82/73.85 & 27.76/81.17  & 26.79/73.72 & 28.71/83.76 & 28.45/87.15 & 28.90/84.24 & {\bf  29.21/85.64} \\ [0.02 cm]
\hline
30 & 25.42/66.20 & 26.63/76.52 & 25.40/66.10  & 27.55/81.89 & 19.86/64.95 & 27.42/82.25 & {\bf 27.67/79.93} \\ [0.02 cm]
\hline
40 & 24.63/61.40 & 25.88/72.64  & 24.61/61.32 & 24.35/73.41 & 16.12/47.77 & 23.51/71.95 & {\bf 26.52/74.72}  \\ [0.02 cm]
\hline
50 & 24.11/58.21 & 25.24/68.98  & 24.10/58.13 & 20.34/57.32 & 14.16/38.63 & 19.48/56.09 & {\bf 25.63/69.82} \\ [0.02 cm]
\hline
60 & 23.75/55.99 & 24.62 /65.58   & 23.74/55.92 & 17.05/41.97 & 12.57/31.87 & 16.59/43.52 & {\bf 24.92/65.56}  \\ [0.02 cm]
\hline
70 & 23.48/54.36 & 24.07/62.37  & 23.48/54.31 & 14.62/30.69 & 11.23/26.50 & 14.53/26.50 & {\bf 24.37/61.81} \\ [0.02 cm]
\hline
80  & 23.27/53.04 & 23.52/59.36  & 23.27/52.99 & 11.65/15.10 & 10.07/22.33 & 11.94/18.98 & {\bf 23.92/58.77} \\ [0.02 cm]
\hline
90 & 23.08/51.99 & 23.01/56.55  & 23.08/51.94 & 08.00/04.10 & 09.04/19.06 & 08.24/05.23 & {\bf 23.53/56.13}  \\  [0.02 cm]
\hline
100  & 22.94/51.13 & 22.53/53.90  & 22.94/51.10  & 06.60/1.23 & 08.12/16.39 & 06.66/01.47 & {\bf 23.24/54.14} \\ [0.02 cm]
\hline 
\end{tabular}
\label{PSNR_SSIM3}
\end{table*}

\begin{table*}[!htp]
\setlength{\tabcolsep}{2pt}
\caption{Comparison of PSNR/SSIM for different natural images \cite{ImgDatabase1,ImgDatabase2} at  $\sigma=50$. The images used are Ba: Barbara, Co: Couple, Ho: House, Bo: Boat, M: Man, Le: Lena,  and Br:
Bridge. \vspace{2mm}} 
\centering
 \begin{tabular}{|c|c|c|c|c|c|c|c|} 
\hline
{Image}  & NLM& NLEM    &  PCA-NLM & Zero-CPW & One-CPW  & Max-CPW & PNLM
\\ [0.2 cm]
\hline
\textit{Ba} & 22.64/68.91  & 24.55/79.24    & 22.63/68.83  & 19.94/64.66  & 14.16/44.40 & 19.26/62.69   & {\bf 25.41/82.01}   \\ [0.02 cm]
\hline
\textit{Co}  & 22.59/56.24  & 23.80/67.56     & 22.58/56.14 & 19.92/59.54 & 14.16/43.17 & 19.27/58.92 & {\bf 24.56/71.09} \\ [0.02 cm]
 \hline
\textit{Ho} & 24.10/69.59  & 25.50/63.20   & 24.09/69.82  & 20.58/26.27  & 14.15/12.49 & 19.67/23.81   & {\bf 27.35/74.00} \\ [0.02 cm]
\hline
\textit{Bo} & 22.98/62.26  & 24.01/70.29  & 22.97/62.19  & 20.06/58.43  & 14.16/40.85 & 19.38/57.22   & {\bf 25.22/74.01} \\ [0.02 cm]
\hline
\textit{Ma} & 23.65/62.20  & 24.88/71.73   & 23.64/62.13  & 20.20/58.92  & 14.16/40.62  & 19.44/57.67  & {\bf 25.51/73.41} \\ [0.02 cm]
\hline
\textit{Le} & 25.22/75.02  & 26.46/79.20    & 25.22/75.00  & 20.58/57.56  & 14.14/35.42   & 19.63/54.28   & {\bf 27.37/82.11} \\ [0.02 cm]
\hline
\textit{Br} & 19.27/30.87 & 20.26/41.66   & 19.26/30.31 & 17.94/35.56 & 14.14/31.58  & 18.10/41.00 & {\bf21.12/47.93} \\ [0.02 cm]
\hline  
\end{tabular}
\label{PSNR_SSIM2}
\end{table*}

\begin{figure}[!htp]
\centering
\subfloat[Clean image (512 $\times$ 512).]{\includegraphics[scale=0.44]{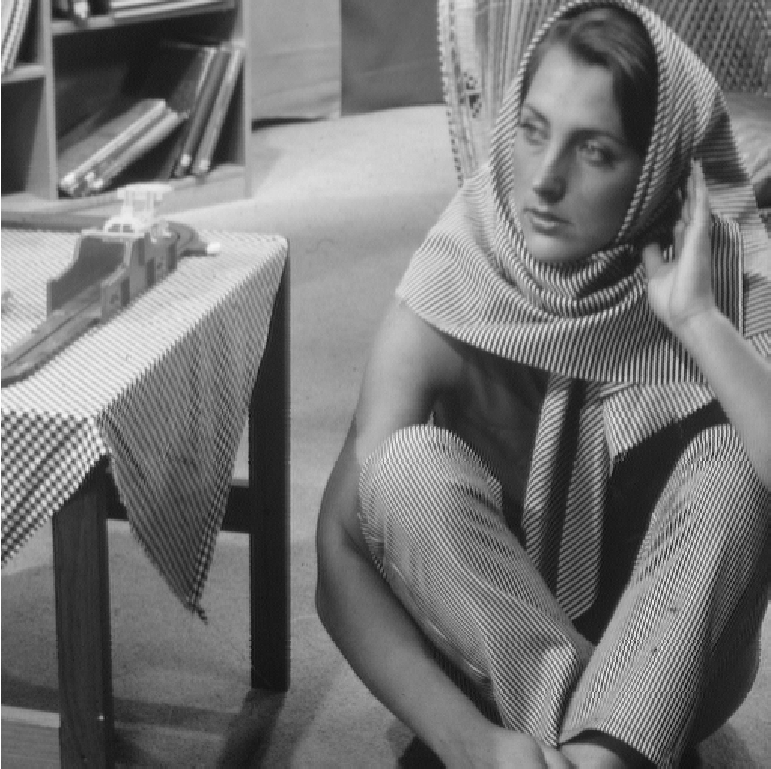}} \hspace{0.1mm}
\subfloat[Noisy image: 16.08 / 52.62.]{\includegraphics[scale=0.44]{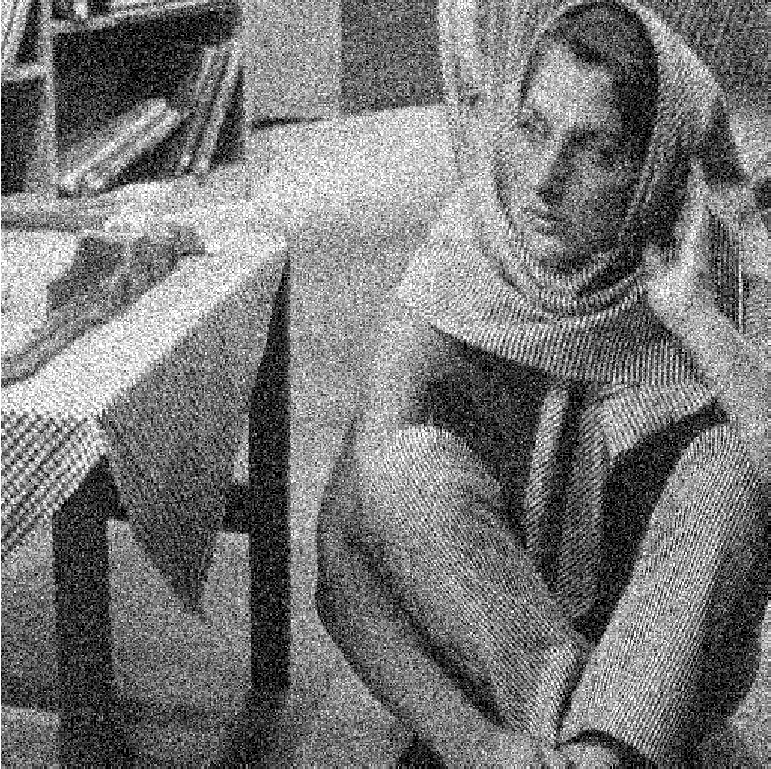}}  \\
\subfloat[NLM: 23.52 / 73.62.]{\includegraphics[scale=0.44]{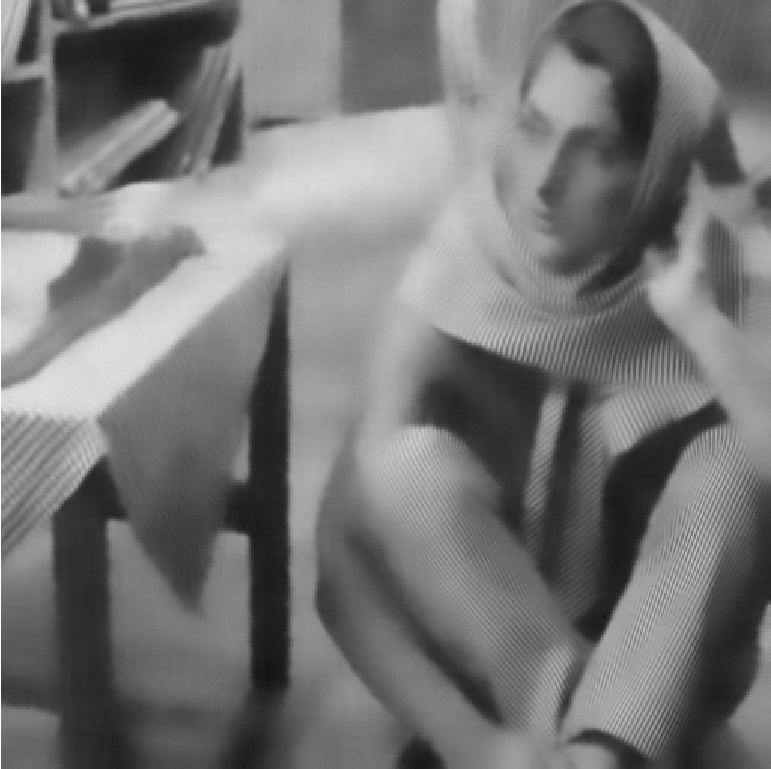}}  \hspace{0.1mm}
\subfloat[PNLM: \textbf{26.45 / 84.10.}] {\includegraphics[scale=0.44]{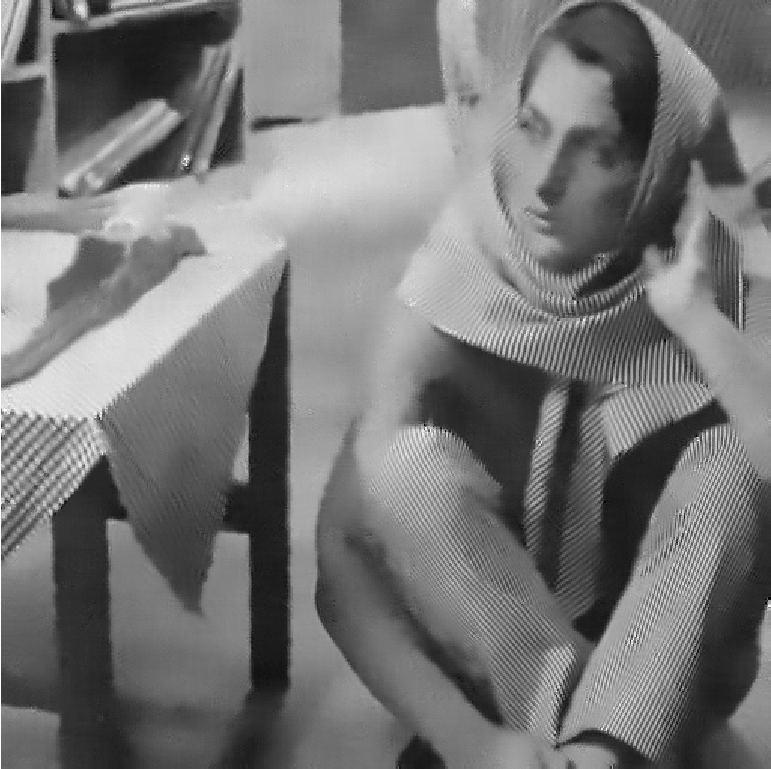}} \\
\subfloat[K-SVD \cite{KSVD2006}: 26.88 / 85.28.]{\includegraphics[scale=0.44]{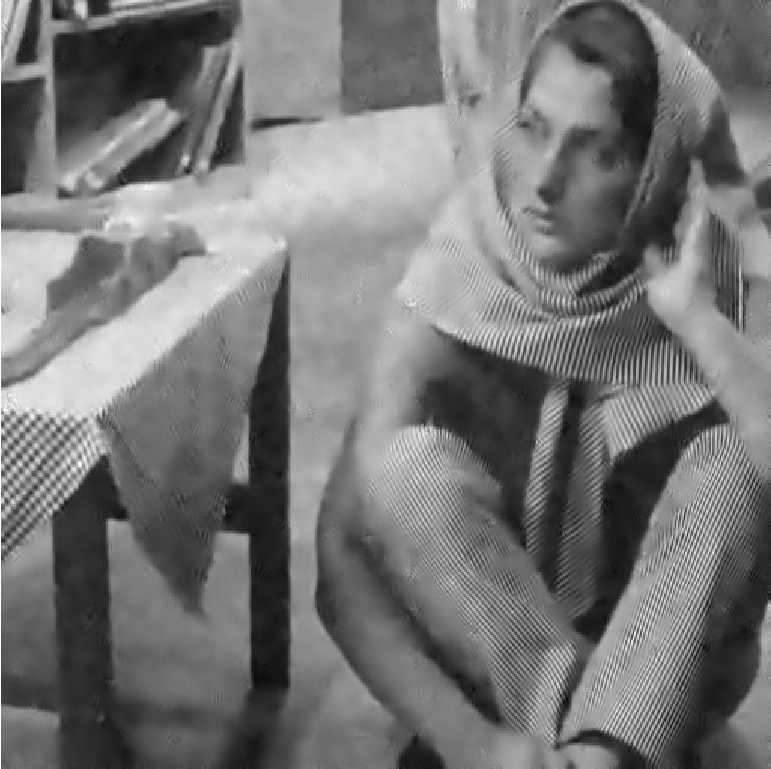}}  \hspace{0.1mm}
\subfloat[BM3D \cite{BM3D2007}: {27.95 / 93.40.}]{\includegraphics[scale=0.44]{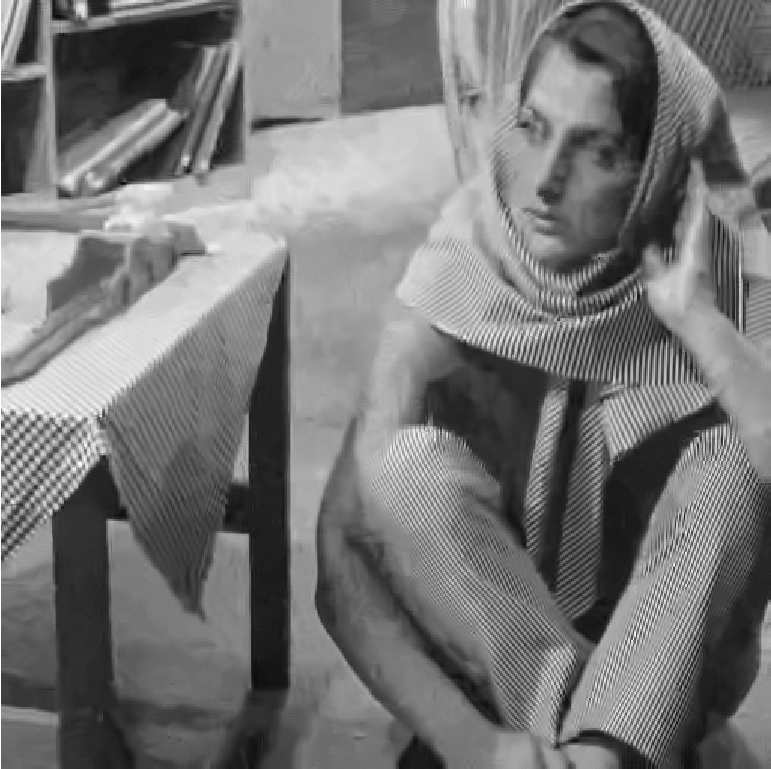}} \\
\caption{Comparison of the denoising results at $\sigma = 40$ for the \textbf{Barbara} image \cite{ImgDatabase1}. The PSNR/SSIM values are shown in the caption.}
\label{visual_2}
\end{figure}


\begin{figure}[!htp]
  \centering
 \subfloat[Clean image (512 $\times$ 512).] {\includegraphics[scale=0.44]{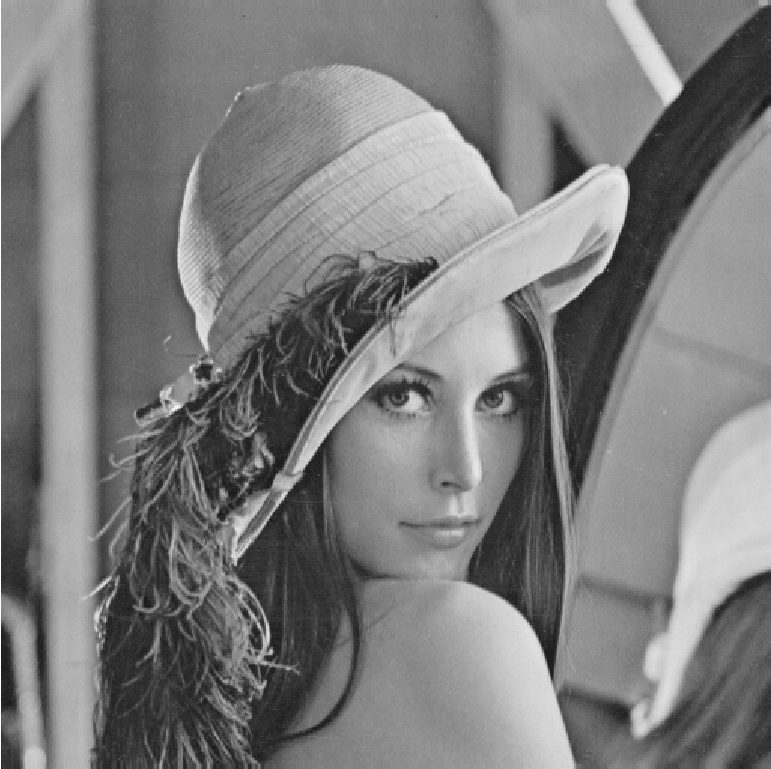}} \hspace{0.1mm}
 \subfloat[Noisy image: 18.60 / 52.90.]{\includegraphics[scale=0.44]{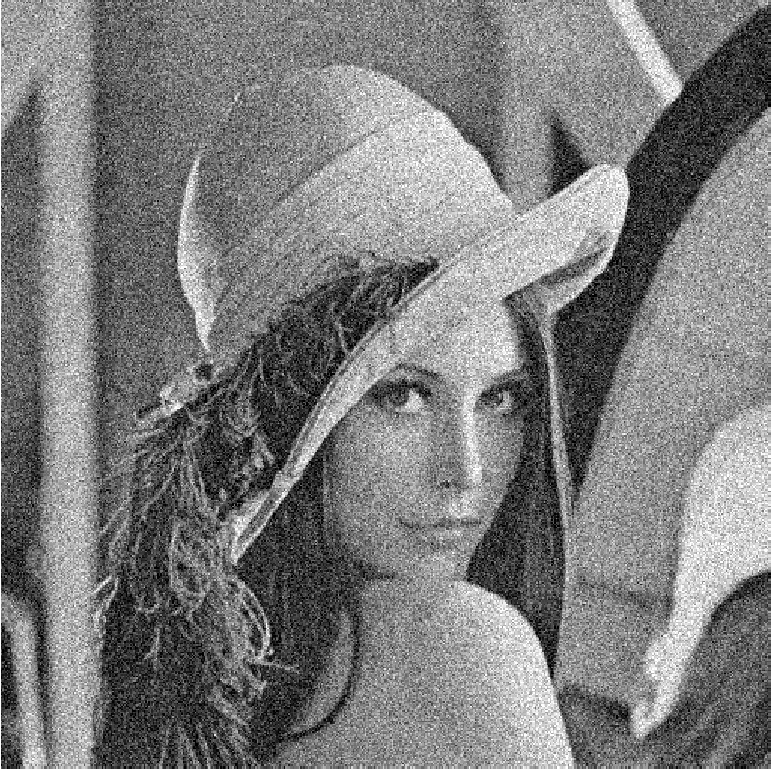}}  \\
\subfloat[NLM: 27.30 / 81.63.]{\includegraphics[scale=0.44]{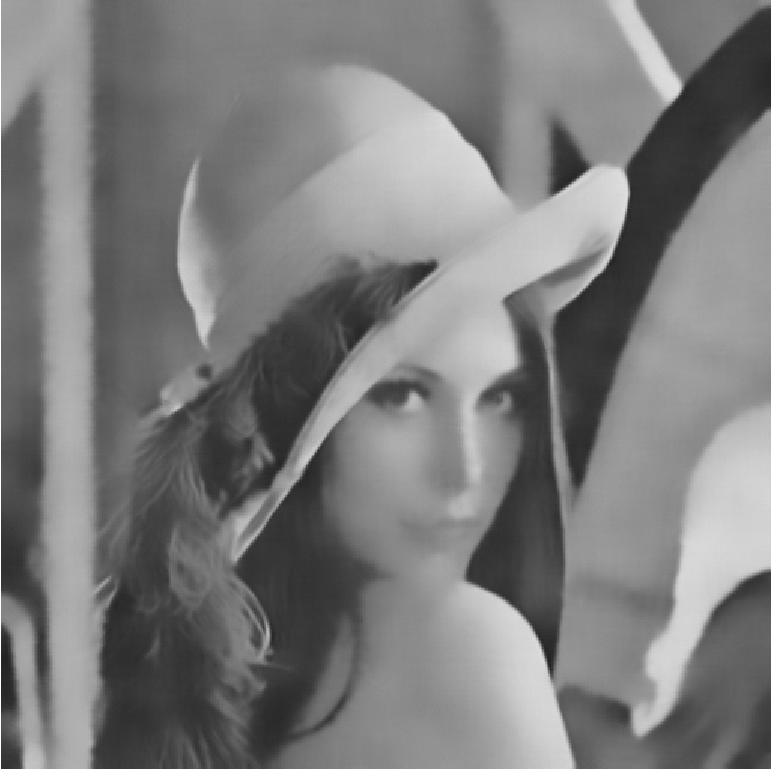}}  \hspace{0.1mm}
\subfloat[PNLM: \textbf{29.70 / 87.61.}] {\includegraphics[scale=0.44]{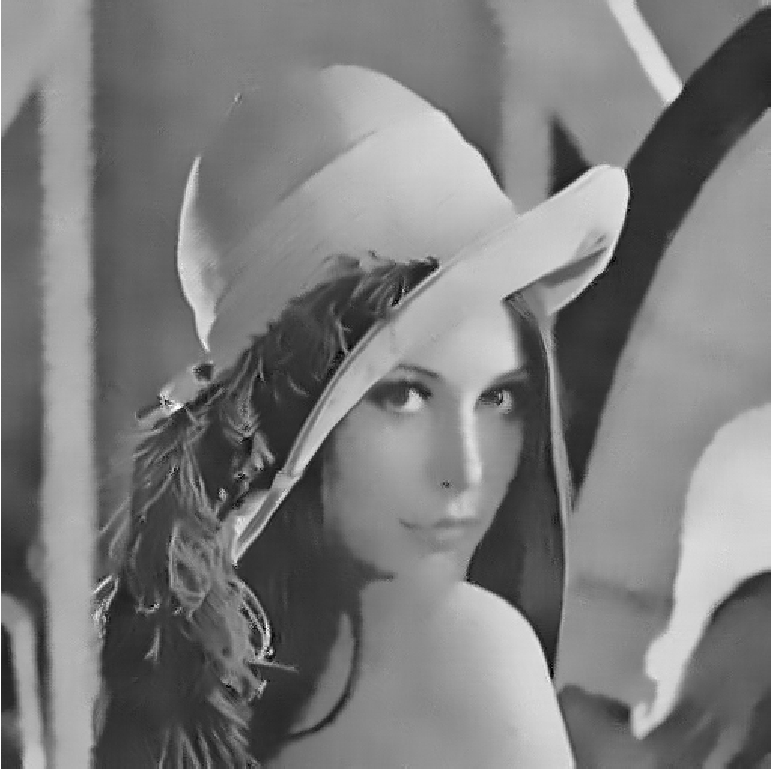}} \\
 \subfloat[K-SVD \cite{KSVD2006}: 30.30 / 89.23.] {\includegraphics[scale=0.44]{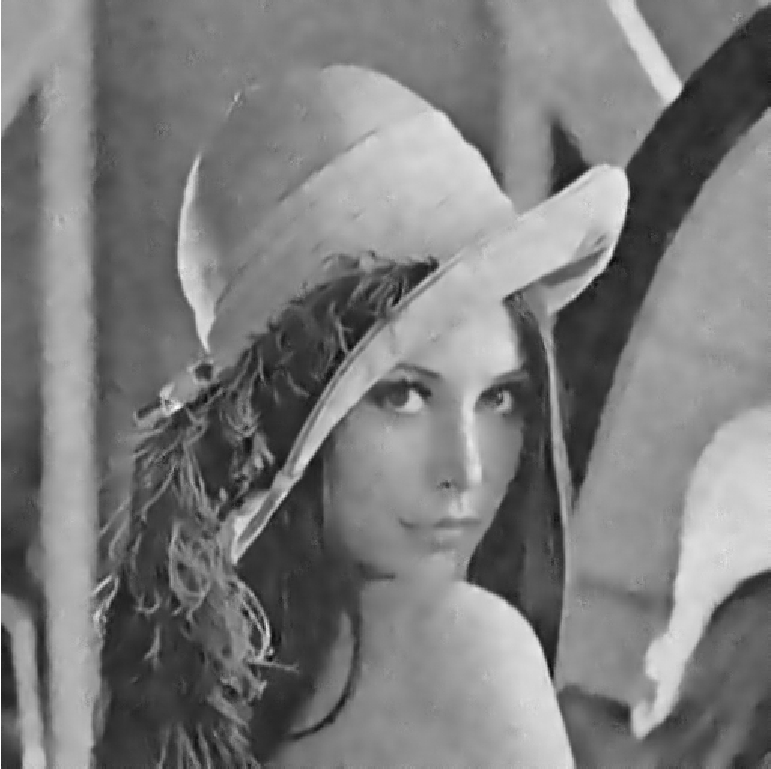}}  \hspace{0.1mm}
 \subfloat[BM3D \cite{BM3D2007}: {31.29 / 95.90.}] {\includegraphics[scale=0.44]{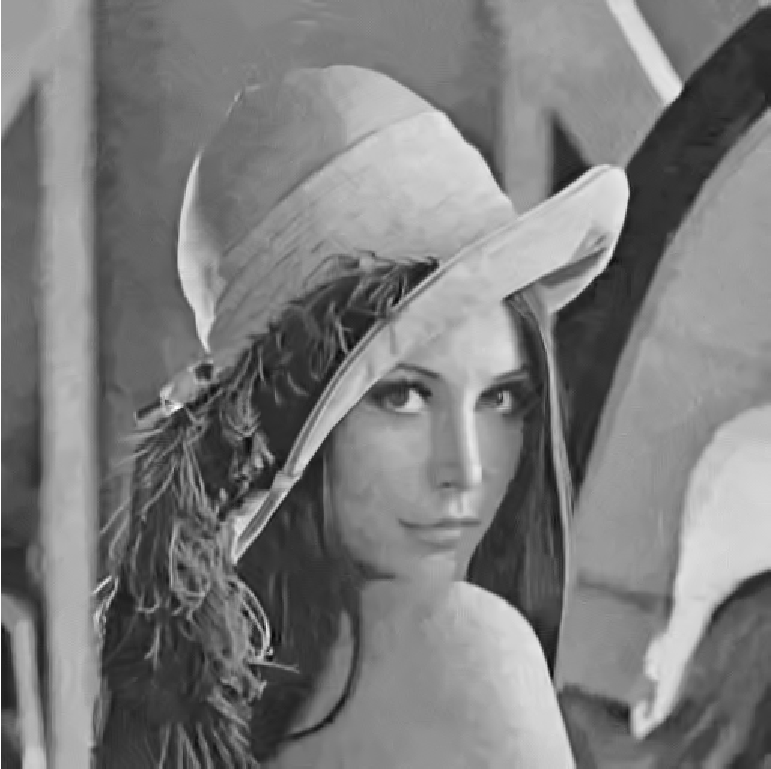}} \\
   \caption{Comparison of the denoising results at $\sigma = 30$ for the \textbf{Lena} image \cite{ImgDatabase1}. The PSNR/SSIM values are shown in the caption.}
    \label{visual}
\end{figure}

For completeness, we have also compared PNLM with the state-of-the-art denoising methods KSVD \cite{KSVD2006} and BM3D \cite{BM3D2007}. A couple of representative results are provided in Figures \ref{visual_2} and \ref{visual}. Notice that the PNLM has successfully reduced the gap in PSNR/SSIM between NLM and K-SVD. This is interesting given that K-SVD is much more complex compared to PNLM. However, there is still a significant margin in the denoising performance with respect to BM3D. In this regard, we wish to make a couple of remarks. First, notice that the denoising using PNLM appears to be somewhat better in homogeneous regions compared to both K-SVD and BM3D. In particular, we see ``ripples'' and ``blocking'' artifacts in the homogeneous regions for the K-SVD and BM3D outputs. In fact, the overall visual quality of the PNLM output appears to be better than the K-SVD output for both  \textit{Lena} and \textit{Barbara}, though this fact is not quite captured by the quality metrics. The other remark is that the BM3D algorithm uses block-matching and aggregation in its processing pipeline. Therefore, the present idea of pruning can in principle be incorporated into BM3D. We wish to investigate this in future work.

\section{Conclusion}

We demonstrated that the denoising capability of NLM can be significantly improved using pruning, where the pruning is controlled using a single global threshold.
The added cost of optimizing the threshold was marginal in comparison to the run-time of NLM.
Moreover, we provided  several denoising results to demonstrate that pruned NLM outperforms several variants of NLM in terms of PSNR and SSIM.
The surprising observation is that, by optimizing a single parameter $\lambda$, we could obtain denoising results that are consistently better than that obtained using sophisticated optimization in \cite{CS2012,SC2013}. It would be interesting to see if the present ideas can be extended to the recently-proposed Vector Nonlocal Euclidean Median \cite{vNLEM}.

\section{Acknowledgements}

The authors wish to thank the reviewers for their useful comments and suggestions. This work was supported by a Startup Grant from the Indian Institute of Science and EMR Grant SERB/F/6047/2016-2017 from the Department of Science and Technology, Government of India.

\section{Appendix}
\label{appendix}
In this section, we provide the main steps in the derivation of  \eqref{sure_pnlm}. First, we write \eqref{pnlm} as 
\begin{equation*}
 \hat{x}_i =  \frac{\sum_{j \in S(i)} \psi(w_{i, j}) \ y_j} { \sum_{j \in S(i)} \psi(w_{i, j}) },
\end{equation*}
where $\psi(t)$ is given by \eqref{temp}. Then, by the quotient rule for derivatives and after some simplification, we obtain
\begin{align}
\label{temp1}
\!W_i  \frac{\partial \hat{x}_i}{\partial y_i} =    \psi(1)+ \!\!\sum_{j \in S(i)} \! \! y_j  \frac{ \partial \psi(w_{i,j}) }{ \partial y_i } - \hat{x}_i \! \! \sum_{j \in S(i)}\!\! \frac{ \partial \psi(w_{i,j}) }{ \partial y_i },
\end{align}
where $W_i$ is given by  \eqref{temp}. Applying chain rule, we have
\begin{equation}
\label{temp2}
 \frac{\partial \psi(w_{i,j}) }{ \partial y_i }  = \psi'(w_{i,j})   \frac{\partial w_{i,j}}{\partial y_i},
\end{equation}
where 
\begin{equation}
\label{temp3}
  \frac{\partial w_{i,j}}{\partial y_i} = \frac{2 w_{i,j}}{h^2} \cdot
  \begin{cases}
     y_j + y_{2i-j} - 2y_i \ & i-j \in \mathcal{P}  \\
    y_j - y_i & i-j \notin \mathcal{P},
  \end{cases}
\end{equation} 
and
\begin{equation}
\label{psi'}
\psi'(t) =  \frac{ \big( 1+ (1 + \alpha t ) e^{- \alpha (t - \lambda)} \big) }{( 1 + e^{ - \alpha (t - \lambda)} )^2}
\end{equation}
Substituting \eqref{temp2} and \eqref{temp3} in \eqref{temp1},
\begin{eqnarray*}
\begin{aligned}
\!W_i  \frac{\partial \hat{x}_i}{\partial y_i} 
= &  \Big\{  \psi(w_{i,i}) + \sum\limits_{j \in S(i)} \psi'(w_{i,j}) (y_j - \hat{x_i})\frac{\partial w_{i,j}}{\partial y_i}\Big\}\\
= &  \Big\{  \psi(w_{i,i}) + \sum\limits_{j \in S(i)} \frac{2 w_{i,j} \ \psi'(w_{i,j}) }{ h^2} (y_j - \hat{x_i}) (y_j - y_i) \\ 
+ & \sum\limits_{k \in \mathcal{P}} \frac{2 w_{i,i+k} \ \psi'(w_{i,i+k}) }{h^2} (y_{i+k} - \hat{x_i}) (y_{i-k} - y_i) \Big\}.
\end{aligned}
\label{eqn:div_l_2}
\end{eqnarray*}

After performing some reductions, we arrive at \eqref{sure_pnlm}:
\begin{eqnarray*}
\begin{aligned}
  \frac{\partial \hat{x}_i}{\partial y_i} 
= \frac{1}{\!W_i} &  \Big\{  \psi(1) + \sum\limits_{j \in S(i)} \frac{2 w_{i,j} \ \psi'(w_{i,j}) }{ h^2} (y_j - \hat{x_i}) (y_j - y_i) \\ 
+ & \sum\limits_{k \in \mathcal{P}} \frac{2 w_{i,i+k} \ \psi'(w_{i,i+k}) }{h^2} (y_{i+k} - \hat{x_i}) (y_{i-k} - y_i) \Big\}.
\end{aligned}
\label{eqn:div_l_2}
\end{eqnarray*}

\end{document}